\documentclass[10pt,journal,compsoc]{IEEEtran}
\usepackage[T1]{fontenc}
\usepackage[utf8]{inputenc}
\usepackage[english]{babel}
\usepackage{amsmath,amssymb,mathtools,amsthm,mathrsfs,bm}
\usepackage{booktabs,array,graphicx}
\usepackage[nocompress]{cite}
\usepackage{url}
\usepackage{xr-hyper}
\usepackage[hidelinks]{hyperref}
\theoremstyle{plain}
\newtheorem{theorem}{Theorem}
\newtheorem{proposition}[theorem]{Proposition}

\theoremstyle{definition}

\theoremstyle{remark}
\newtheorem{remark}[theorem]{Remark}
\hypersetup{pdftitle={Decision-Target Reversal under Informative Label Missingness},pdfauthor={Fariborz Setoudehtazangi and Geoffrey J. McLachlan}}
\begin{document}
\title{Decision-Target Reversal under Informative Label Missingness}
\author{Fariborz~Setoudehtazangi and Geoffrey~J.~McLachlan%
\IEEEcompsocitemizethanks{%
\IEEEcompsocthanksitem F. Setoudehtazangi is with the Dipartimento di
Scienze Statistiche, Universit\`a di Padova, Padova, Italy.
\IEEEcompsocthanksitem G. J. McLachlan is with the School of Mathematics
and Physics, The University of Queensland, Brisbane, Australia.
Corresponding author: Geoffrey J. McLachlan
(e-mail: \href{mailto:g.mclachlan@uq.edu.au}{g.mclachlan@uq.edu.au}).}}
\IEEEtitleabstractindextext{\begin{abstract}
Informative label missingness can improve classification efficiency because
the pattern of label availability may itself contain information about the
discriminant structure. We show that this advantage need not persist across
deployment objectives: with the data-generating model and statistical
experiments held fixed, informative partial classification can be favourable
under one loss and unfavourable under another. We call this phenomenon
decision-target reversal. For binary Gaussian discrimination, we derive a
local cost-sensitive excess-risk expansion and, in one dimension,
obtain explicit critical cost ratios at which the first-order risk
ordering reverses. We show that the
required covariance geometry arises under finite uncertainty-dependent
missingness mechanisms with a fixed marginal missing-label proportion. We
then extend the analysis to multiclass finite-action problems through
loss-specific active-boundary curvature, yielding a general
decision-target frontier. Monte Carlo experiments illustrate the predicted
reversals, and a descriptive gastrointestinal-lesion application with
agreement-defined label availability illustrates sensitivity to deployment
costs.
The results show that favourable informative missingness is a property of
the statistical experiment relative to the deployment decision, rather
than of the experiment alone.
\end{abstract}
\begin{IEEEkeywords}Informative label missingness, partially classified data, decision-target reversal, cost-sensitive classification, classification risk, comparison of experiments.\end{IEEEkeywords}}
\maketitle
\IEEEdisplaynontitleabstractindextext
\section{Introduction}
\label{sec:introduction}

Partially classified samples arise when feature measurements are observed
for all training units but class labels are available for only a subset.
This setting is common when definitive classification requires expert
assessment, laboratory testing, follow-up, or another procedure that is
more difficult or costly than obtaining the features themselves. In
likelihood-based semi-supervised classification, observations without class
labels can still contribute through the marginal feature distribution, but
the unavailable labels ordinarily represent a loss of information about the
discriminant structure.

This usual efficiency ordering can change when label availability is
informative. Following the missing-data framework of
Rubin~\cite{rubin1976inference}, Ahfock and McLachlan~\cite{ahfock2020paradox} considered Gaussian
discrimination in which the probability that a class label is missing
depends on classification uncertainty. Because the missingness mechanism
shares parameters with the classifier, the observed pattern of label
availability can itself carry information about the discriminant model.
The information supplied by the missingness indicators may then compensate,
partly or more than fully, for the information lost through unavailable
class labels. Consequently, an informatively partially classified sample
can have smaller asymptotic classification risk than the corresponding
ordinary completely classified sample.

Throughout this paper, complete classification (CC) refers to the ordinary
fully labelled experiment based on $(\boldsymbol Y,Z)$, without treating a
mechanism-generated missingness indicator as an additional observation.
This distinction is important. If both all class labels and the missingness
indicator were observed, giving the augmented experiment
$(\boldsymbol Y,Z,M)$, the partially classified observation would be a
coarsening of that experiment and could not dominate it. The favourable
missingness phenomenon concerns comparison with the ordinary CC benchmark.

Subsequent work has clarified the information structure underlying this
phenomenon and extended it beyond the original Gaussian setting.
Information decompositions separate the conditional class-label information
lost through missingness from the information contributed by the observed
missingness indicators \cite{wang2026learning}. Related analyses have
established favourable-missingness effects in other parametric
classification models \cite{zhou2026exponential,wu2026weibull}. More generally, Setoudehtazangi and McLachlan~\cite{SetoudehtazangiMcLachlan2026a} developed a
nuisance-adjusted information decomposition together with the local
second-order geometry of multiclass zero--one classification risk. This
analysis builds on the classical excess-risk approach to discriminant
efficiency, in which the expected increase in classification error due to
estimation of the discriminant rule admits an $O(n^{-1})$ expansion
\cite{efron1975efficiency}. In the informative-missingness setting,
favourable performance does not require improvement in every parameter
direction: classification efficiency depends on how the resulting
estimation uncertainty projects onto directions that perturb the active
Bayes decision boundary.

These results compare statistical experiments under a fixed classification
objective. At deployment, different errors may carry different losses,
and changing those losses changes the Bayes action even when the
data-generating distribution remains fixed
\cite{elkan2001foundations}. Dependence of classifier comparisons
on misclassification costs is well established.
Adams and Hand~\cite{adams1999comparing} studied comparisons under uncertain costs,
and Drummond and Holte~\cite{drummond2006cost} developed cost curves for comparing
classifiers across operating conditions. Under correct posterior
specification, a fixed fitted model can also be deployed under
different costs by changing its posterior threshold
\cite{dmochowski2010maximum}.

Cost-sensitive learning with incompletely labelled samples has been
studied through semi-supervised support vector machines and EM-based
procedures \cite{li2010cost,WangEtAl2012}. Our contribution concerns
the first-order comparison of complete classification and
classifier-linked informative partial classification: we derive
explicit deployment-cost crossings and prove that their required
covariance geometry is generated by a finite missingness mechanism
at a fixed marginal missing-label proportion.

We study precisely this question. The data-generating model, the
informative missingness mechanism, and the two statistical experiments are
held fixed. Only the deployment loss is varied. We show that the resulting
classification-risk ordering need not be invariant: the same informative
partially classified experiment can be favourable for one deployment target
and unfavourable for another. We refer to this phenomenon as
\emph{decision-target reversal}.

The possibility of target-dependent ordering is consistent with the
classical theory for comparison of statistical experiments
\cite{blackwell1951comparison,blackwell1953equivalent,
torgersen1991comparison}. Blackwell dominance gives a uniform risk ordering
over decision problems; in its absence, a pair of experiments need not
retain the same ordering under different losses. Our aim is not to restate this general decision-theoretic fact, but to
identify the classification-specific mechanism that produces reversal
under informative label missingness, to characterize explicitly the
deployment losses at which the ordering changes, and to determine when
the covariance geometry required for such a reversal can arise from the
missingness mechanism itself.
The mechanism is transparent in binary Gaussian discrimination. Let
\[
\boldsymbol D
=
\boldsymbol V_{\mathrm{PC}}
-
\boldsymbol V_{\mathrm{CC}}
\]
denote the difference between the asymptotic covariance matrices of the
classifier parameters under informative partial and complete
classification. Informative missingness can reduce estimation uncertainty
in some parameter directions while increasing it in others, so
$\boldsymbol D$ may be indefinite. For a binary cost matrix
$\boldsymbol C$, let $t_C$ denote the corresponding Bayes decision
boundary. We show that the leading difference in deployment risk is
governed by
\[
Q_{\boldsymbol D}(t_C)
=
d_{00}+2t_Cd_{01}+t_C^2d_{11}
+\sum_{j=2}^{p}d_{jj}.
\]
Thus the statistical experiment determines the fixed covariance geometry
$\boldsymbol D$, whereas the deployment loss determines which part of that
geometry is relevant to classification by moving the Bayes boundary.
Changing the loss can therefore change the sign of the risk comparison
without changing either training experiment.

In one dimension, this representation yields explicit critical cost
ratios at which the ordering reverses. We
also show that the required indefinite covariance geometry is not an
artificial assumption: for a finite uncertainty-dependent missingness
mechanism with a fixed marginal missing-label proportion, it occurs over a
nonempty open region of the parameter space. The phenomenon is therefore
structural rather than confined to a specially chosen numerical example.

The risk comparison is not specific to informative label missingness:
it applies to any two regular estimators of the same model.
Informative missingness provides the concrete likelihood construction
for which we establish the covariance geometry required for reversal.
The same principle extends to multiclass finite-action classification. For
a loss matrix $\boldsymbol C$, let $\boldsymbol H_C$ denote the local
curvature of the corresponding population Bayes risk. Under regularity
conditions, an estimator with asymptotic covariance $\boldsymbol V$
satisfies
\[
E\{R_C(\widehat{\boldsymbol\theta})\}-R_C^*
=
\frac{1}{2n}
\operatorname{tr}
(\boldsymbol H_C\boldsymbol V)
+
o(n^{-1}).
\]
For informative partial versus complete classification, the leading
comparison is therefore
\[
\Delta(C)
=
\operatorname{tr}
\left\{
\boldsymbol H_C
(\boldsymbol V_{\mathrm{PC}}-\boldsymbol V_{\mathrm{CC}})
\right\}.
\]
This yields the decision-target frontier
\[
\mathcal B_{\boldsymbol D}
=
\left\{
\boldsymbol C:
\operatorname{tr}
(\boldsymbol H_C\boldsymbol D)=0
\right\}.
\]
The general formulation also reveals a distinction from the one-dimensional binary case:
indefiniteness of $\boldsymbol D$ alone does not guarantee reversal,
because the family of curvature matrices generated by admissible losses
need not probe both favourable and unfavourable directions of
$\boldsymbol D$.

The paper therefore characterizes how deployment costs alter the
first-order comparison of complete and informative partial classification.
We derive the local
cost-sensitive excess-risk expansion for binary Gaussian discrimination
and explicit reversal thresholds in one dimension; establish that the covariance geometry
required for reversal is generated by finite informative-missingness
mechanisms at a fixed marginal missing-label proportion; and extend the
analysis to multiclass finite-action problems through loss-specific active
decision-boundary curvature. Monte Carlo experiments illustrate the
predicted reversal at finite sample sizes, including under unequal class
probabilities. A descriptive gastrointestinal-lesion application examines
cost sensitivity when label availability is determined by expert agreement.
The application treats the relative error costs as sensitivity parameters,
rather than as estimates of clinical utility.

The remainder of the paper is organized as follows.
Section~\ref{sec:decision-risk} develops the loss-specific local risk
geometry for binary classification and characterizes decision-target
reversal. Section~\ref{sec:informative-missingness} derives the efficient
information under informative label missingness and establishes the
existence of the required covariance geometry.
Section~\ref{sec:general-loss} gives the multiclass finite-action
extension. Section~\ref{sec:simulation} presents the population and
finite-sample numerical results, Section~\ref{sec:gastro} gives the
gastrointestinal-lesion application, and Section~\ref{sec:discussion}
concludes.
\section{Decision-target-dependent classification risk}
\label{sec:decision-risk}

Let $Z\in\{1,2\}$ denote the class label and
$\boldsymbol Y\in\mathbb R^p$ the feature vector. We begin with the
canonical homoscedastic Gaussian discrimination model
\begin{equation}
\begin{aligned}
\boldsymbol Y\mid Z=1 &\sim N_p\left(\frac{\Delta}{2}\boldsymbol e_1,\boldsymbol I_p\right),\\
\boldsymbol Y\mid Z=2 &\sim N_p\left(-\frac{\Delta}{2}\boldsymbol e_1,\boldsymbol I_p\right).
\end{aligned}
\label{eq:canonical-model}
\end{equation}
where $\Delta>0$ and
\[
\Pr(Z=1)=\pi_1,
\qquad
\Pr(Z=2)=\pi_2=1-\pi_1.
\]
Writing
\[
\lambda=\log\left(\frac{\pi_1}{\pi_2}\right),
\]
the posterior log-odds at the true parameter value are
\begin{equation}
\eta_0(\boldsymbol y)
=
\log
\frac{\Pr(Z=1\mid\boldsymbol Y=\boldsymbol y)}
     {\Pr(Z=2\mid\boldsymbol Y=\boldsymbol y)}
=
\lambda+\Delta y_1.
\label{eq:true-log-odds}
\end{equation}

Suppose that the consequences of the two types of
misclassification are not necessarily equal. Let $C_{12}>0$ denote
the loss incurred by assigning an observation from class 2 to class 1,
and let $C_{21}>0$ denote the loss incurred by assigning an observation
from class 1 to class 2. Correct classifications have zero loss. The
Bayes rule assigns an observation to class 1 whenever
\[
C_{12}\Pr(Z=2\mid\boldsymbol Y=\boldsymbol y)
\leq
C_{21}\Pr(Z=1\mid\boldsymbol Y=\boldsymbol y),
\]
or, equivalently,
\begin{equation}
\eta_0(\boldsymbol y)\geq r,
\qquad
r=\log\left(\frac{C_{12}}{C_{21}}\right).
\label{eq:cost-rule}
\end{equation}
Its decision boundary is therefore
\begin{equation}
y_1=t_C,
\qquad
t_C=\frac{r-\lambda}{\Delta}.
\label{eq:cost-boundary}
\end{equation}
Thus changing the relative misclassification losses moves the
deployment boundary even though the underlying statistical model is
unchanged.

Consider a local perturbation of the discriminant,
\begin{equation}
\eta_{\boldsymbol h}(\boldsymbol y)
=
\lambda+h_0
+
(\Delta+h_1)y_1
+
\sum_{j=2}^p h_jy_j,
\label{eq:local-discriminant}
\end{equation}
where $\boldsymbol h=(h_0,\ldots,h_p)^\top$ is close to zero. Let
$R_C(\boldsymbol h)$ denote the expected deployment loss of the
classifier induced by \eqref{eq:local-discriminant}, evaluated under
the true distribution, and let $R_C^*=R_C(\boldsymbol 0)$ denote the
Bayes risk.

The following result gives the local cost-sensitive analogue of the
classical excess-error expansion for normal discrimination.

\begin{proposition}
\label{prop:binary-risk-expansion}
Under model \eqref{eq:canonical-model}, as
$\|\boldsymbol h\|\to0$,
\begin{equation}
R_C(\boldsymbol h)-R_C^*
=
\frac{m_C}{2\Delta}
\left\{
(h_0+t_Ch_1)^2
+
\sum_{j=2}^p h_j^2
\right\}
+
o(\|\boldsymbol h\|^2),
\label{eq:binary-risk-expansion}
\end{equation}
where
\begin{equation}
m_C
=
C_{21}\pi_1
\phi\left(t_C-\frac{\Delta}{2}\right)
=
C_{12}\pi_2
\phi\left(t_C+\frac{\Delta}{2}\right)
>0.
\label{eq:mC}
\end{equation}
\end{proposition}

The proof is given in Supplementary
Section~\ref{sec:proof-binary-risk}.

To connect the local expansion with estimation uncertainty, parameterize
the fitted discriminant by
\[
\eta_{\boldsymbol\beta}(\boldsymbol y)
=
\lambda+\beta_0+\beta_1y_1
+\sum_{j=2}^{p}\beta_jy_j,
\qquad
\boldsymbol\beta^\star
=
(0,\Delta,0,\ldots,0)^\top.
\]
Thus the perturbation in \eqref{eq:local-discriminant} is
$\boldsymbol h=\boldsymbol\beta-\boldsymbol\beta^\star$.
The constant $\lambda$ centers the true intercept for the analysis;
the fitted intercept $\lambda+\widehat\beta_0$ is estimated, and the
true class probabilities are not supplied as known quantities.
Suppose that an estimator $\widehat{\boldsymbol\beta}$ in these
discriminant coordinates satisfies
\[
\sqrt n
(\widehat{\boldsymbol\beta}-\boldsymbol\beta^\star)
\overset{d}{\longrightarrow}
N_{p+1}(\boldsymbol 0,\boldsymbol V),
\]
together with the moment conditions required to average the local
quadratic expansion. Proposition~\ref{prop:binary-risk-expansion}
then yields
\begin{equation}
\begin{aligned}
&E\{R_C(\widehat{\boldsymbol\beta})\}-R_C^*\\
&\quad=\frac{m_C}{2\Delta n}
\left\{v_{00}+2t_Cv_{01}+t_C^2v_{11}
+\sum_{j=2}^p v_{jj}\right\}+o(n^{-1}),
\end{aligned}
\label{eq:expected-cost-risk}
\end{equation}
where $\boldsymbol V=(v_{jk})$.

\begin{remark}[Connection with the classical excess-error expansion]
For equal misclassification costs, $C_{12}=C_{21}$, we have
$r=0$ and hence
\[
t_C=-\frac{\lambda}{\Delta}.
\]
Consequently, the covariance contribution in
\eqref{eq:expected-cost-risk} reduces to
\[
v_{00}
-\frac{2\lambda}{\Delta}v_{01}
+\frac{\lambda^2}{\Delta^2}v_{11}
+\sum_{j=2}^p v_{jj},
\]
which is the covariance combination appearing in the classical
excess-error expansion of Efron~\cite{efron1975efficiency} and subsequently
used by Ahfock and McLachlan~\cite{ahfock2020paradox} for the ordinary zero--one
classification comparison. Thus the equal-cost case is a single point
in the loss-indexed family considered here. Varying the deployment loss
moves the Bayes decision boundary and thereby changes the covariance
projection relevant to classification.
\end{remark}

Consider two statistical experiments with asymptotic covariance
matrices $\boldsymbol V_1$ and $\boldsymbol V_2$, and write
\[
\boldsymbol D=\boldsymbol V_1-\boldsymbol V_2
=(d_{jk}).
\]
Since $m_C/(2\Delta)>0$, their first-order deployment-risk ordering is
determined by
\begin{equation}
Q_{\boldsymbol D}(t_C)
=
d_{00}
+
2t_Cd_{01}
+
t_C^2d_{11}
+
\sum_{j=2}^p d_{jj}.
\label{eq:target-comparison}
\end{equation}
The statistical experiment is fixed in
\eqref{eq:target-comparison}; only the deployment loss, through
$t_C$, changes. Consequently, an efficiency ordering obtained under
one decision target need not persist under another.

For $p=1$, this dependence can be characterized exactly.
\begin{theorem}
\label{thm:binary-reversal}
Let $p=1$ and define
\[
Q_{\boldsymbol D}(t)
=
d_{00}+2d_{01}t+d_{11}t^2.
\]

If $d_{11}\neq0$, define
\[
\mathcal D_{\boldsymbol D}
=
d_{01}^2-d_{00}d_{11}.
\]
When $\mathcal D_{\boldsymbol D}>0$, the first-order
deployment-risk ordering changes at the two boundary locations
\[
t^{(1)}
=
\frac{-d_{01}-\sqrt{\mathcal D_{\boldsymbol D}}}{d_{11}},
\qquad
t^{(2)}
=
\frac{-d_{01}+\sqrt{\mathcal D_{\boldsymbol D}}}{d_{11}}.
\]
The corresponding positive cost ratios are
\[
\widetilde c_j
=
\exp\{\lambda+\Delta t^{(j)}\},
\qquad j=1,2.
\]
Writing
\[
c_{\rm low}
=
\min(\widetilde c_1,\widetilde c_2),
\qquad
c_{\rm high}
=
\max(\widetilde c_1,\widetilde c_2),
\]
gives
\[
0<c_{\rm low}<c_{\rm high}<\infty.
\]
When $\mathcal D_{\boldsymbol D}=0$, the quadratic is tangent to zero
and there is no genuine sign reversal, whereas
$\mathcal D_{\boldsymbol D}<0$ yields no real critical boundary
location.

If $d_{11}=0$ and $d_{01}\neq0$, there is a single critical boundary
location
\[
t_*=-\frac{d_{00}}{2d_{01}},
\]
and hence a single positive critical cost ratio
\[
c_*=\exp\{\lambda+\Delta t_*\}.
\]
If $d_{11}=d_{01}=0$, the first-order comparison is independent of the
decision target.

In particular, if $\boldsymbol D$ is indefinite, then
\[
\mathcal D_{\boldsymbol D}
=
-\det(\boldsymbol D)>0,
\]
and there exist positive cost ratios for which the two experiments
have opposite first-order deployment-risk orderings.
\end{theorem}

The proof is given in Supplementary Section~\ref{sec:proof-binary-reversal}.

Theorem~\ref{thm:binary-reversal} separates two questions that are
identical under a fixed loss but distinct when the deployment
objective varies. The covariance difference $\boldsymbol D$
describes how the statistical experiment changes estimation
uncertainty, whereas the cost ratio determines which directions of
that uncertainty perturb the relevant decision boundary. An
indefinite $\boldsymbol D$ therefore does not assign an intrinsic
``better'' experiment: the answer depends on the decision target.
\section{Informative label missingness}
\label{sec:informative-missingness}

Let $M$ denote the label-missingness indicator, with $M=1$ when the
class label is unavailable. We assume
\begin{equation}
M\perp Z\mid\boldsymbol Y,
\label{eq:mar-assumption}
\end{equation}
and allow the probability of a missing label to depend on the
classifier through
\begin{equation}
\Pr(M=1\mid\boldsymbol Y=\boldsymbol y)
=
q(\boldsymbol y;\boldsymbol\beta,\boldsymbol\xi),
\label{eq:missingness-general}
\end{equation}
where $\boldsymbol\beta$ denotes the parameters governing
classification and $\boldsymbol\xi$ contains parameters specific to
the missingness mechanism. Thus the missingness indicator may itself
carry information about $\boldsymbol\beta$.

We compare two experiments. Under complete classification,
\[
\mathcal E_{\mathrm{CC}}:
\qquad
(\boldsymbol Y,Z)
\]
is observed. Under partial classification,
\[
\mathcal E_{\mathrm{PC}}:
\qquad
\{\boldsymbol Y,M,(1-M)Z\}
\]
is observed. These are distinct statistical experiments: in
$\mathcal E_{\mathrm{PC}}$, the missingness indicator is an observed,
parameter-linked random variable, whereas complete classification
refers to direct availability of the class label.

For the information identities below, $\boldsymbol\beta$ indexes
the full distribution of $(\boldsymbol Y,Z)$, and
$\boldsymbol\xi$ is the only additional nuisance parameter.
Additional unknown location or covariance parameters require the
full generative information calculation and subsequent nuisance
adjustment or projection into discriminant coordinates.

Let $\boldsymbol I_Y$ denote the Fisher information for
$\boldsymbol\beta$ in the marginal distribution of $\boldsymbol Y$,
and let
\[
\boldsymbol I_{Z\mid Y}(\boldsymbol y)
=
E\left[
\boldsymbol S_{Z\mid Y}
\boldsymbol S_{Z\mid Y}^{\top}
\mid
\boldsymbol Y=\boldsymbol y
\right]
\]
denote the conditional information in the class label. The information
under complete classification is
\begin{equation}
\boldsymbol I_{\mathrm{CC}}
=
\boldsymbol I_Y
+
E\{\boldsymbol I_{Z\mid Y}(\boldsymbol Y)\}.
\label{eq:icc-decomposition}
\end{equation}

Under partial classification, an unavailable label removes its
conditional label information, but the missingness event contributes
information whenever $q$ depends on $\boldsymbol\beta$. After
eliminating the nuisance parameter $\boldsymbol\xi$, the efficient
information is
\begin{equation}
\boldsymbol I_{\mathrm{PC}}^{\mathrm{eff}}
=
\boldsymbol I_{\mathrm{CC}}
-
E\left[
q(\boldsymbol Y)
\boldsymbol I_{Z\mid Y}(\boldsymbol Y)
\right]
+
\boldsymbol I_M^{\mathrm{eff}},
\label{eq:ipc-general}
\end{equation}
where
\begin{equation}
\boldsymbol I_M^{\mathrm{eff}}
=
\boldsymbol I_{\beta\beta}^{M}
-
\boldsymbol I_{\beta\xi}^{M}
(\boldsymbol I_{\xi\xi}^{M})^{-1}
\boldsymbol I_{\xi\beta}^{M}.
\label{eq:missing-efficient-information}
\end{equation}
The two terms on the right-hand side of
\eqref{eq:ipc-general} have different roles. The first is the
information lost through unavailable class labels; the second is the
classifier information retained in the pattern of label availability
after nuisance variation in the missingness mechanism has been
removed.

More generally, if the generative parameter is $\boldsymbol\theta$
and the discriminant coefficients are
$\boldsymbol\beta=\boldsymbol\psi(\boldsymbol\theta)$, apply
\eqref{eq:ipc-general}--\eqref{eq:missing-efficient-information}
in the full $\boldsymbol\theta$ coordinates. If
$\boldsymbol K_{\mathrm{PC}}$ is the resulting information after
eliminating $\boldsymbol\xi$, then
\[
\begin{aligned}
\boldsymbol V_{\mathrm{PC}}
&=\boldsymbol G_0\boldsymbol K_{\mathrm{PC}}^{-1}\boldsymbol G_0^\top,\\
\boldsymbol V_{\mathrm{CC}}
&=\boldsymbol G_0(\boldsymbol I_{\mathrm{CC}}^\theta)^{-1}\boldsymbol G_0^\top,\\
\boldsymbol G_0
&=\left.\frac{\partial\boldsymbol\psi(\boldsymbol\theta)}
{\partial\boldsymbol\theta^\top}\right|_{\boldsymbol\theta_0}.
\end{aligned}
\]
These covariance matrices follow from nuisance elimination
and the delta method; see
Setoudehtazangi and McLachlan~\cite{SetoudehtazangiMcLachlan2026a}.

To expose the resulting geometry, consider the two-parameter
Gaussian likelihood
\begin{equation}
\begin{aligned}
\Pr(Z=1)&=\operatorname{expit}(b),\\
Y\mid Z=1&\sim N(\delta/2,1),\\
Y\mid Z=2&\sim N(-\delta/2,1).
\end{aligned}
\label{eq:restricted-fitted-model}
\end{equation}
where $b\in\mathbb R$ and $\delta>0$ are estimated. The midpoint of
the two means and the within-class variance are known and fixed at
zero and one, respectively. These are restrictions on the fitted
likelihood, not consequences of standardizing the true distribution.
The fitted posterior log-odds are $b+\delta y$.

The symmetric configuration has true values $b=0$ and $\delta=\Delta$,
so that $\pi_1=\pi_2=1/2$ and $\eta_0(y)=\Delta y$.
Equal class probabilities are not imposed during fitting.
At this configuration $(\beta_0,\beta_1)=(b,\delta)$.
For an unequal-prior configuration, the centered coordinates in
Section~\ref{sec:decision-risk} are
$(\beta_0,\beta_1)=(b-\lambda,\delta)$.
Both parameters of the mechanism
$q(y;b,\delta,\boldsymbol\xi)
=\operatorname{expit}\{\xi_0+\xi_1(b+\delta y)^2\}$
are also estimated freely.

We use the uncertainty-dependent mechanism
\begin{equation}
q(y)
=
\operatorname{expit}
\left\{
\xi_0+\xi_1\eta_0(y)^2
\right\}
=
\operatorname{expit}
\left\{
\xi_0+\xi_1\Delta^2y^2
\right\}.
\label{eq:uncertainty-missingness}
\end{equation}
For $\xi_1<0$, labels are more likely to be unavailable near the Bayes
decision boundary, where the posterior evidence for the two classes is
weakest.

Let
\[
\tau(y)=\Pr(Z=1\mid Y=y)
       =\operatorname{expit}(\Delta y).
\]
In the discriminant parameterization introduced in
Section~\ref{sec:decision-risk},
\[
\boldsymbol\beta=(\beta_0,\beta_1)^\top,
\qquad
\boldsymbol\beta^\star=(0,\Delta)^\top,
\]
the conditional label information is
\begin{equation}
\boldsymbol I_{Z\mid Y}(y)
=
\tau(y)\{1-\tau(y)\}
\begin{pmatrix}
1 & y\\
y & y^2
\end{pmatrix}.
\label{eq:conditional-label-information}
\end{equation}

The next result shows that informative missingness affects the two
discriminant directions differently.

\begin{proposition}
\label{prop:symmetric-information}
Under the fitted likelihood \eqref{eq:restricted-fitted-model},
at the true values $b=0$ and $\delta=\Delta$, and under the
missingness mechanism \eqref{eq:uncertainty-missingness}, the efficient information
contributed by the missingness indicator is
\begin{equation}
\boldsymbol I_M^{\mathrm{eff}}
=
\begin{pmatrix}
4\xi_1^2\Delta^2
E\!\left[
q(Y)\{1-q(Y)\}Y^2
\right]
&
0\\[1mm]
0&0
\end{pmatrix}.
\label{eq:IM-symmetric}
\end{equation}
Moreover,
\begin{equation}
\boldsymbol I_{\mathrm{CC}}
=
\frac14\boldsymbol I_2
\label{eq:ICC-symmetric}
\end{equation}
and
\begin{equation}
\boldsymbol I_{\mathrm{PC}}^{\mathrm{eff}}
=
\begin{pmatrix}
J_0&0\\
0&J_1
\end{pmatrix},
\label{eq:IPC-symmetric}
\end{equation}
where
\begin{align}
J_0
&=
\frac14
-
E\!\left[
q(Y)\tau(Y)\{1-\tau(Y)\}
\right]\notag\\
&\quad+
4\xi_1^2\Delta^2
E\!\left[
q(Y)\{1-q(Y)\}Y^2
\right],
\label{eq:J0}\\
J_1
&=
\frac14
-
E\!\left[
q(Y)\tau(Y)\{1-\tau(Y)\}Y^2
\right].
\label{eq:J1}
\end{align}
\end{proposition}

The proof is given in Supplementary Section~\ref{sec:proof-symmetric-information}.

Since the expectation subtracted in \eqref{eq:J1} is strictly
positive whenever labels are missing with positive probability,
\[
J_1<\frac14.
\]
Thus partial classification necessarily loses information in the
slope direction in this symmetric model. The intercept direction is
different. It gains information relative to complete classification
precisely when
\begin{equation}
4\xi_1^2\Delta^2
E\!\left[
q(Y)\{1-q(Y)\}Y^2
\right]
>
E\!\left[
q(Y)\tau(Y)\{1-\tau(Y)\}
\right].
\label{eq:indefinite-condition}
\end{equation}
When \eqref{eq:indefinite-condition} holds,
\[
\boldsymbol D
=
(\boldsymbol I_{\mathrm{PC}}^{\mathrm{eff}})^{-1}
-
\boldsymbol I_{\mathrm{CC}}^{-1}
\]
has one negative and one positive eigenvalue. The partially classified
experiment then improves estimation in one discriminant direction
while degrading it in another. By
Theorem~\ref{thm:binary-reversal}, neither experiment has a uniform
first-order advantage over all deployment cost ratios.

\subsection{Existence of a target-sensitive regime}
\label{subsec:existence}

Condition~\eqref{eq:indefinite-condition} identifies the geometry
required for decision-target reversal, but does not by itself establish
that such a configuration can arise at a fixed amount of missingness.
We now show that it does.

Write
\[
\xi_1=-a,
\qquad a>0,
\]
and, for each $a$, choose $\xi_0=\xi_0(a)$ so that
\begin{equation}
E\{q_a(Y)\}=\gamma,
\qquad 0<\gamma<1,
\label{eq:fixed-gamma}
\end{equation}
where
\begin{equation}
q_a(y)
=
\operatorname{expit}
\left\{
\xi_0(a)-a\Delta^2y^2
\right\}.
\label{eq:qa}
\end{equation}
The calibration in \eqref{eq:fixed-gamma} is used only to compare
data-generating mechanisms having the same marginal proportion of
missing labels. The parameters $\xi_0$ and $\xi_1$ remain unrestricted
nuisance parameters in the statistical model.

Let
\[
\boldsymbol V_{\mathrm{PC}}(a)
=
\left\{
\boldsymbol I_{\mathrm{PC}}^{\mathrm{eff}}(a)
\right\}^{-1},
\qquad
\boldsymbol V_{\mathrm{CC}}
=
\boldsymbol I_{\mathrm{CC}}^{-1},
\]
and define
\[
\boldsymbol D(a)
=
\boldsymbol V_{\mathrm{PC}}(a)
-
\boldsymbol V_{\mathrm{CC}}.
\]

\begin{theorem}
\label{thm:existence-reversal}
Under the fitted likelihood \eqref{eq:restricted-fitted-model},
suppose that the true values are $b=0$ and $\delta=\Delta>0$,
and that the missingness mechanism is given by \eqref{eq:qa}, with
$\xi_0(a)$ calibrated according to \eqref{eq:fixed-gamma}.
For every $\gamma\in(0,1)$, there exists a finite
$a_0=a_0(\Delta,\gamma)$ such that, for all $a>a_0$,
\[
\boldsymbol D(a)
=
\boldsymbol V_{\mathrm{PC}}(a)
-
\boldsymbol V_{\mathrm{CC}}
\]
is indefinite. More specifically,
\begin{equation}
D_{00}(a)<0,
\qquad
D_{11}(a)>0.
\label{eq:D-signs}
\end{equation}
Consequently, there exist positive misclassification-cost ratios for
which partial classification has smaller first-order deployment risk
than complete classification and other cost ratios for which the
ordering is reversed.
\end{theorem}

The proof is given in Supplementary Section~\ref{sec:proof-existence-reversal}.

Theorem~\ref{thm:existence-reversal} shows that target dependence is
not merely a possibility created by an arbitrary indefinite covariance
matrix. It arises from the informative-missingness model itself while
the marginal proportion of missing labels is held fixed. In the
symmetric model, the missingness indicator can contribute sufficient
information about the intercept direction to offset the information
lost through unavailable labels, while no corresponding efficient
missingness information remains in the slope direction.

The result is established through a limiting argument as $a$ becomes
large, but the conclusion concerns finite values of $a$. Moreover, whenever the inequalities in \eqref{eq:D-signs} are strict
at a finite parameter value, continuity of the information matrices
implies that they persist in an ordinary neighborhood of that
configuration. If the fixed-$\gamma$ calibration is imposed, the
implicit-function theorem shows that the same inequalities persist
in a relative neighborhood within the calibrated family. Thus the
target-sensitive regime is not confined to an isolated parameter
configuration.
\section{Extension to multiclass finite-action losses}
\label{sec:general-loss}

The one-dimensional binary Gaussian model provides explicit
decision-target reversal thresholds. We now consider multiclass
problems with finite action losses. Building on the zero--one
active-boundary expansion of
Setoudehtazangi and McLachlan~\cite{SetoudehtazangiMcLachlan2026a}, we derive the loss-specific
curvature matrix and characterize reversal through the family
of curvature matrices attainable by varying the deployment loss.

Let $Z\in\{1,\ldots,K\}$ and let
$\tau_k(\boldsymbol y;\boldsymbol\theta)$ denote the posterior
probability of class $k$. Suppose that the available actions are
$a\in\{1,\ldots,A\}$ and that
$\boldsymbol C=(C_{ak})$ is a finite loss matrix. The conditional
risk of action $a$ is
\begin{equation}
\rho_a(\boldsymbol y;\boldsymbol\theta)
=
\sum_{k=1}^K
C_{ak}\tau_k(\boldsymbol y;\boldsymbol\theta),
\label{eq:conditional-action-risk}
\end{equation}
and the Bayes action minimizes \eqref{eq:conditional-action-risk}.

The decision problem is invariant to multiplication of
$\boldsymbol C$ by a common positive constant and to addition of the
same class-dependent vector to every action row, since these
transformations preserve the ordering of the conditional action
risks. Accordingly, the decision-target frontier below is
understood on the resulting equivalence classes of loss matrices.

For two actions $a$ and $b$, define the contrast
\begin{equation}
g_{ab}(\boldsymbol y;\boldsymbol\theta)
=
\rho_a(\boldsymbol y;\boldsymbol\theta)
-
\rho_b(\boldsymbol y;\boldsymbol\theta).
\label{eq:action-risk-contrast}
\end{equation}
At the true parameter $\boldsymbol\theta_0$, let
$\mathcal F_{ab}^{C}$ denote the active decision face on which
\[
\rho_a(\boldsymbol y;\boldsymbol\theta_0)
=
\rho_b(\boldsymbol y;\boldsymbol\theta_0)
<
\rho_c(\boldsymbol y;\boldsymbol\theta_0),
\qquad
c\notin\{a,b\}.
\]
Assume that the feature distribution has fixed support
$\mathcal S\subseteq\mathbb R^p$ and a continuous, locally bounded
density, and that the contrasts $g_{ab}$ are jointly twice
continuously differentiable in $(\boldsymbol y,\boldsymbol\theta)$
near $\boldsymbol\theta_0$. The minimizing action is unique almost
everywhere. On compact subsets of the interior of $\mathcal S$, the
active tie sets have a finite regular stratification: pairwise faces
are smooth with nonzero spatial contrast gradient, and higher-order
active intersections are transverse and have codimension at least
two. Assume that compact sets $K_L$ exhaust $\mathcal S$, with
boundaries transverse to these strata. The local chart and
codimension conditions are detailed in Supplementary
Section~\ref{sec:proof-general-loss}.

Define
\begin{equation}
\boldsymbol b_{ab}(\boldsymbol y)
=\nabla_{\boldsymbol\theta}
 g_{ab}(\boldsymbol y;\boldsymbol\theta_0).
\label{eq:b-ab}
\end{equation}
Assume the surface-integrability condition
\begin{equation}
\sum_{a<b}\int_{\mathcal F_{ab}^{C}}
\frac{f_{\boldsymbol Y}(\boldsymbol y;\boldsymbol\theta_0)
      \|\boldsymbol b_{ab}(\boldsymbol y)\|^2}
     {\|\nabla_{\boldsymbol y}g_{ab}
          (\boldsymbol y;\boldsymbol\theta_0)\|}
\,dS(\boldsymbol y)<\infty.
\label{eq:general-loss-surface-integrability}
\end{equation}
Let $a_{\boldsymbol h}(\boldsymbol y)$ minimize
$\rho_a(\boldsymbol y;\boldsymbol\theta_0+\boldsymbol h)$ under a
fixed tie convention, and set
\[
\mathcal L_C(\boldsymbol y;\boldsymbol h)
=\rho_{a_{\boldsymbol h}(\boldsymbol y)}
       (\boldsymbol y;\boldsymbol\theta_0)
 -\min_a\rho_a(\boldsymbol y;\boldsymbol\theta_0).
\]
In addition, assume the uniform quadratic tail condition
\begin{equation}
\lim_{L\to\infty}\limsup_{\boldsymbol h\to\boldsymbol 0}
\frac{1}{\|\boldsymbol h\|^2}
\int_{\mathcal S\setminus K_L}
 f_{\boldsymbol Y}(\boldsymbol y;\boldsymbol\theta_0)
 \mathcal L_C(\boldsymbol y;\boldsymbol h)\,d\boldsymbol y=0.
\label{eq:general-loss-quadratic-tail}
\end{equation}

\begin{proposition}
\label{prop:general-loss-expansion}
Under the preceding regularity conditions, as
$\|\boldsymbol h\|\to0$,
\begin{equation}
R_C(\boldsymbol\theta_0+\boldsymbol h)-R_C^*
=
\frac12
\boldsymbol h^\top
\boldsymbol H_C
\boldsymbol h
+
o(\|\boldsymbol h\|^2),
\label{eq:general-loss-expansion}
\end{equation}
where
\begin{equation}
\boldsymbol H_C
=
\sum_{a<b}
\int_{\mathcal F_{ab}^{C}}
\frac{
f_{\boldsymbol Y}(\boldsymbol y;\boldsymbol\theta_0)
\,
\boldsymbol b_{ab}(\boldsymbol y)
\boldsymbol b_{ab}(\boldsymbol y)^\top
}{
\left\|
\nabla_{\boldsymbol y}
g_{ab}(\boldsymbol y;\boldsymbol\theta_0)
\right\|
}
\,dS(\boldsymbol y).
\label{eq:HC-general}
\end{equation}
If, in addition,
\[
\sqrt n
(\widehat{\boldsymbol\theta}-\boldsymbol\theta_0)
\overset{d}{\longrightarrow}
N(\boldsymbol 0,\boldsymbol V),
\]
the sequence
\[
\left\{
\left\|
\sqrt n
(\widehat{\boldsymbol\theta}-\boldsymbol\theta_0)
\right\|^2
\right\}_{n\geq1}
\]
is uniformly integrable, and the remainder in
\eqref{eq:general-loss-expansion}, evaluated at
$\boldsymbol h=\widehat{\boldsymbol\theta}-\boldsymbol\theta_0$,
is uniformly integrable after multiplication by $n$, then
\begin{equation}
E\{R_C(\widehat{\boldsymbol\theta})\}-R_C^*
=
\frac{1}{2n}
\operatorname{tr}
(\boldsymbol H_C\boldsymbol V)
+
o(n^{-1}).
\label{eq:general-loss-expected-risk}
\end{equation}
\end{proposition}

The proof is given in Supplementary
Section~\ref{sec:proof-general-loss}.

For the complete- and partially classified experiments, write
\begin{equation}
\boldsymbol D
=
\boldsymbol V_{\mathrm{PC}}
-
\boldsymbol V_{\mathrm{CC}}.
\label{eq:D-general}
\end{equation}
Their first-order deployment-risk difference is then governed by
\begin{equation}
\Delta_C
=
\operatorname{tr}
(\boldsymbol H_C\boldsymbol D).
\label{eq:delta-C}
\end{equation}
With the convention in \eqref{eq:D-general},
\[
\Delta_C<0
\]
means that partial classification has smaller first-order deployment
risk, whereas $\Delta_C>0$ favors complete classification.

The set
\begin{equation}
\mathcal B_D
=
\left\{
\boldsymbol C:
\operatorname{tr}
(\boldsymbol H_C\boldsymbol D)=0
\right\}
\label{eq:favourability-frontier}
\end{equation}
defines the decision-target frontier. In contrast with the
one-dimensional binary Gaussian case, this frontier is generally not available in closed
form because changing $\boldsymbol C$ moves the active decision faces
and may also change which actions are locally optimal. Decision-target
reversal occurs whenever the attainable family
$\{\boldsymbol H_C\}$ intersects both sides of the hyperplane
\[
\boldsymbol H
\mapsto
\operatorname{tr}(\boldsymbol H\boldsymbol D)=0.
\]

More generally, for a class $\mathcal C$ of admissible deployment
losses, decision-target reversal occurs precisely when
\begin{equation}
\inf_{\boldsymbol C\in\mathcal C}
\operatorname{tr}
(\boldsymbol H_C\boldsymbol D)
<0
<
\sup_{\boldsymbol C\in\mathcal C}
\operatorname{tr}
(\boldsymbol H_C\boldsymbol D).
\label{eq:attainable-curvature-criterion}
\end{equation}
Thus indefiniteness of $\boldsymbol D$ alone is not sufficient in the
general multiclass problem: the relevant question is whether the
attainable family of loss-specific curvature matrices
$\{\boldsymbol H_C:\boldsymbol C\in\mathcal C\}$ probes both favourable
and unfavourable directions of the covariance difference.

\section{Finite-sample validation of decision-target reversal}
\label{sec:simulation}

We next examine whether the decision-target reversal predicted by the
asymptotic theory is visible at practically relevant sample sizes, and
whether the local risk expansion explains the finite-sample ordering of
the two experiments.  The purpose of the experiment is to compare deployment risks and
examine how closely the local quadratic expansion tracks the
estimated finite-sample risk curves.
Differences from the population curve can reflect finite-sample
estimation effects as well as Monte Carlo variability.

We use the symmetric Gaussian configuration
\[
\pi_1=\pi_2=\frac12,
\qquad
\Delta=1.5,
\]
with missing-label probability
\[
q(y)
=
\operatorname{expit}
\left\{
\xi_0+\xi_1(\Delta y)^2
\right\}.
\]
The mechanism strength is fixed at
\[
\xi_1=-0.25,
\]
and the intercept is calibrated to
\[
\xi_0=-0.206813,
\]
which gives the marginal missing-label proportion
\[
\Pr(M=1)=0.30.
\]
For this configuration, the population covariance difference derived
in Section~\ref{sec:informative-missingness} is
\[
\boldsymbol D
=
\boldsymbol V_{\mathrm{PC}}
-
\boldsymbol V_{\mathrm{CC}}
=
\begin{pmatrix}
-0.471312 & 0\\
0 & 0.447128
\end{pmatrix}.
\]
Thus the partially classified experiment is asymptotically more
efficient in the intercept direction and less efficient in the slope
direction.  The corresponding population critical cost ratios are
\[
c_{\rm low} = 0.214374,\qquad
c_{\rm high} = 4.664740.
\]
Consequently, the first-order theory predicts that partial
classification is favourable for deployment targets between these
critical values and unfavourable for sufficiently asymmetric costs
outside this interval.

For each
\[
n\in\{500,1000,2000\},
\]
we generated \(5000\) independent samples.  For every sample we fitted
both the complete-classification estimator and the partially classified
maximum-likelihood estimator, including estimation of the missingness
parameters.  The same generated sample was used for the two estimators
within each replication.  For deployment-risk evaluation, we normalized $C_{21}=1$ and
set $C_{12}=c$. Risk was evaluated exactly under the
data-generating distribution over a grid of positive cost ratios.  All \(15000\) paired fits were successful.  Across
all replications and sample sizes, the realized mean missing-label
proportion was \(0.29994\), close to its target value of \(0.30\).

For a given cost ratio \(c\), let
\[
\widehat{\Delta R}_n(c)
=
\frac{1}{B}
\sum_{b=1}^{B}
\left\{
R_C(\widehat{\boldsymbol\beta}_{\mathrm{PC}}^{(b)})
-
R_C(\widehat{\boldsymbol\beta}_{\mathrm{CC}}^{(b)})
\right\},
\qquad B=5000,
\]
denote the Monte Carlo estimate of the difference in deployment risk.
Negative values therefore favour partial classification.  We compare
this quantity with the first-order approximation
\[
\begin{aligned}
\Delta R_n^{(1)}(c;\boldsymbol D)
&=\frac{m_C}{2\Delta n}
\{d_{00}+2t_Cd_{01}+t_C^2d_{11}\},\\
t_C&=\frac{\log c-\lambda}{\Delta},
\end{aligned}
\]
first using the population covariance difference \(\boldsymbol D\), and
then using the empirical covariance difference
\[
\widehat{\boldsymbol D}_n
=
\widehat{\operatorname{Cov}}
\left\{
\sqrt n
(\widehat{\boldsymbol\beta}_{\mathrm{PC}}
-\boldsymbol\beta^\star)
\right\}
-
\widehat{\operatorname{Cov}}
\left\{
\sqrt n
(\widehat{\boldsymbol\beta}_{\mathrm{CC}}
-\boldsymbol\beta^\star)
\right\}.
\]
This is a diagnostic comparison based on the same paired
replications as the Monte Carlo risk curve. Agreement between
these summaries does not independently establish crossing
precision or separate finite-sample effects from Monte Carlo
variability. We also computed empirical second-moment differences
about the true parameter. Replacing the centered covariance
differences by these second-moment differences changes the
matrices by approximately $0.08\%$--$0.11\%$ in relative
Frobenius norm across the four simulation configurations.
Empirical estimation bias therefore contributes little to
the reported quadratic comparisons.

We also compared covariance-based approximations with risk curves
computed from disjoint batches of simulation replications.
The symmetric equal-cost advantage persists in both
risk-evaluation batches, whereas the unequal-prior advantage
remains unresolved.
The split comparison and its Monte Carlo limitations are reported
in Supplementary Section~\ref{sec:supp-independent-batches}.

\begin{figure*}[t]
    \centering
    \includegraphics[width=0.78\textwidth]{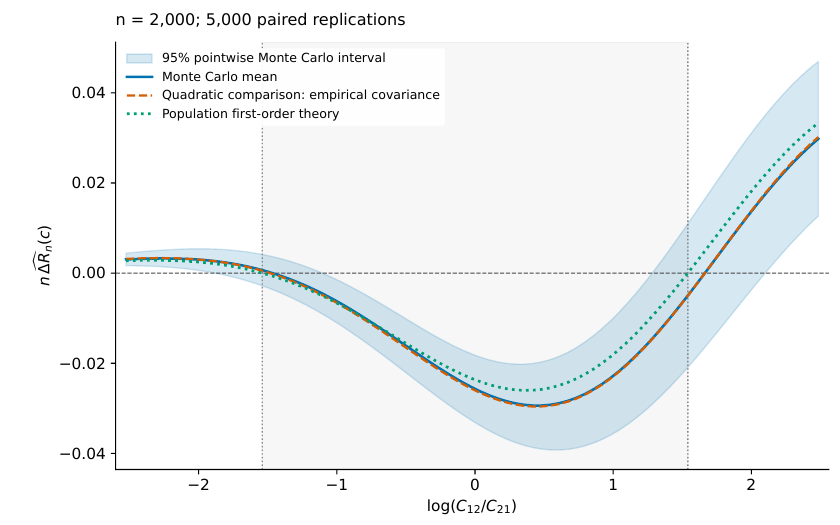}
    \caption{
    Finite-sample decision-target reversal for
    $\pi_1=\pi_2=1/2$, $\Delta=1.5$,
    $\Pr(M=1)=0.30$, and $\xi_1=-0.25$, based on
    5000 paired replications at $n=2000$, with $C_{21}=1$.
    The solid curve estimates $n\widehat{\Delta R}_n(c)$;
    the dashed curve is the quadratic comparison using the
    empirical covariance difference $\widehat{\boldsymbol D}_n$;
    and the dotted curve is the population first-order approximation.
    The blue ribbon gives pointwise 95\% Monte Carlo intervals
    calculated from paired risk differences; it is not a
    simultaneous band.
    The grey region and vertical lines indicate the population
    favourable interval, bounded by $c_{\rm low}=0.214374$
    and $c_{\rm high}=4.664740$.
    Negative values favour partial classification and positive
    values favour complete classification.}
    \label{fig:finite-sample-reversal}
\end{figure*}

Figure~\ref{fig:finite-sample-reversal} shows the resulting scaled risk
difference
\[
n\widehat{\Delta R}_n(c)
\]
for \(n=2000\), together with the two first-order approximations.  The
estimated finite-sample curve exhibits the sign pattern predicted by
the theory: partial classification is favourable over an intermediate
range of deployment costs, whereas complete classification is
preferable in the outer portions of the displayed range. The curve
is closely tracked by the quadratic approximation using
$\widehat{\boldsymbol D}_n$. This agreement is consistent with the
covariance--risk relation underlying the asymptotic calculation.

The agreement can be quantified through the zero crossings of the risk
difference.  Table~\ref{tab:critical-costs} reports the empirical
crossings obtained directly from the Monte Carlo risk curve, the
crossings implied by \(\widehat{\boldsymbol D}_n\), and the population
critical values.  At every sample size, the crossings of the observed
risk curve are closely reproduced by those obtained from the empirical
covariance geometry.  For example, at \(n=2000\), the Monte Carlo
crossings are \(0.2310\) and \(5.2777\), while the corresponding
first-order values based on \(\widehat{\boldsymbol D}_{2000}\) are
\(0.2269\) and \(5.2774\).  This agreement is a diagnostic of the quadratic risk representation,
since both estimates use the same replications.
It does not establish independent predictive accuracy or precise
crossing locations.

\begin{table*}[t]
\centering
\caption{
Critical cost ratios obtained from the Monte Carlo deployment-risk
curve, from the first-order approximation using the empirical
covariance difference \(\widehat{\boldsymbol D}_n\), and from the
population theory.  The simulation uses \(5000\) replications for each
sample size. The Monte Carlo crossings are obtained by linear interpolation
of the mean risk difference in $\log c$ on the evaluated cost grid.
}
\label{tab:critical-costs}
\begin{tabular}{c cc cc cc}
\hline
& \multicolumn{2}{c}{Monte Carlo risk}
& \multicolumn{2}{c}{Empirical \(\widehat{\boldsymbol D}_n\)}
& \multicolumn{2}{c}{Population theory} \\
\(n\) &
$c_{\rm low}$ & $c_{\rm high}$ &
$c_{\rm low}$ & $c_{\rm high}$ &
$c_{\rm low}$ & $c_{\rm high}$ \\
\hline
500
& 0.2928 & 4.2418
& 0.2874 & 4.6125
& 0.2144 & 4.6647 \\
1000
& 0.3043 & 3.3520
& 0.2993 & 3.3785
& 0.2144 & 4.6647 \\
2000
& 0.2310 & 5.2777
& 0.2269 & 5.2774
& 0.2144 & 4.6647 \\
\hline
\end{tabular}
\end{table*}

The same conclusion is obtained from a global comparison of the risk
curves.  The root mean squared discrepancy between the Monte Carlo risk
difference and the population first-order approximation is
\(6.12\times10^{-6}\), \(9.03\times10^{-6}\), and
\(1.52\times10^{-6}\) for \(n=500,1000,\) and \(2000\),
respectively.  When the population covariance difference is replaced
by \(\widehat{\boldsymbol D}_n\), the corresponding discrepancies are
\(2.70\times10^{-6}\), \(2.65\times10^{-7}\), and
\(1.19\times10^{-7}\).  The substantially closer agreement with the
empirical-covariance approximation supports the local quadratic
description of the reported risk curves. The discrepancy from the
population curve may include both finite-sample effects and Monte
Carlo variability in the estimated covariance difference.

As a simple interior target, consider equal misclassification costs,
\(c=1\).  The population theory predicts that partial classification is
favourable at this target.  The Monte Carlo risk differences are
\[
-4.03\times10^{-5},\qquad
-1.80\times10^{-5},\qquad
-1.28\times10^{-5}
\]
for \(n=500,1000,\) and \(2000\), respectively.  Their corresponding
95\% Monte Carlo intervals are
\[
\begin{gathered}
(-5.56,-2.51)\times10^{-5},\\
(-2.53,-1.08)\times10^{-5},\\
(-1.66,-0.91)\times10^{-5},
\end{gathered}
\]
all lying below zero.  The first-order population values are,
respectively,
\[
-4.73\times10^{-5},\qquad
-2.37\times10^{-5},\qquad
-1.18\times10^{-5}.
\]
Thus the favourable ordering at equal costs is resolved at the
pointwise 95\% Monte Carlo level for every sample size considered.

The symmetric simulation resolves opposite risk orderings at
separated deployment targets. Crossing locations nevertheless
have appreciable Monte Carlo uncertainty. At $n=2000$, we
resampled the 5000 paired replication IDs 2000 times, recomputed
the mean risk curve, and interpolated its crossings in $\log c$.
All resamples produced two crossings within the evaluated range.
Their 2.5th and 97.5th percentile ranges are
$(0.155,0.331)$ and $(3.635,8.036)$, respectively.
Both contain the corresponding population critical values.
The observed crossing displacements should therefore not be
attributed solely to finite-sample effects.

As a robustness check, we considered unequal class probabilities,
$\pi_1=0.30$ and $\pi_2=0.70$. The population covariance difference
has a substantial off-diagonal component and remains indefinite,
with eigenvalues $-0.3085$ and $0.4123$. This establishes that
reversal is attainable beyond the equal-prior configuration.
The finite-sample mean curve has two estimated sign changes,
but its favourable portion is not resolved from zero at the
present Monte Carlo precision. The population calculation
therefore supplies the robustness argument; the finite-sample
curve remains illustrative. Full results are reported in
Supplementary Section~\ref{sec:supp-unequal-prior}.

\section{Gastrointestinal-lesion application}
\label{sec:gastro}

We illustrate decision-target dependence using the gastrointestinal-lesion
data of Mesejo and Pizarro~\cite{mesejo2016gastrointestinal} and the partially classified
construction of Lyu et al.~\cite{LyuEtAl2024}. The data comprise $76$ lesions,
with histology providing the reference class: $55$ require resection
and $21$ do not. We retain the four previously selected image-derived
features from the published analysis. The comparison is conditional
on this historical representation and does not evaluate the original
feature-selection procedure.

Agreement among seven endoscopists determines label availability.
When all seven agree, the histological class label is available for
fitting; otherwise it is masked. This gives $35$ labelled and $41$
unlabelled lesions. Expert agreement determines only label availability.
Histology supplies every available training label and every
evaluation label, including cases in which a unanimous expert
decision disagrees with histology.

We compare complete classification (CC) with informative partial
classification (FULL) using two Gaussian classes with unrestricted
means and a shared covariance matrix. This shared-covariance
specification replaces the component-specific covariance specification
of the published analysis. Under FULL, label availability is modelled
jointly through a logistic function of log posterior classification
entropy, with both missingness parameters estimated.

Evaluation uses leave-one-out fitting. In each fold, feature
standardization and all starting values are computed from the
$75$ training lesions. CC uses all training histological labels,
whereas FULL uses only the agreement-defined available labels.
The evaluated lesion contributes neither to preprocessing nor to
initialization or fitting. CC is fitted by its exact Gaussian
maximum-likelihood solution; FULL is fitted from multiple training-only
starts. All $76$ paired fits satisfy the numerical checks described
in Supplementary Section~\ref{sec:supp-gastro}. These checks establish reproducible
numerical local optima, rather than a guarantee of global maximization.

\begin{figure*}[t]
    \centering
    \includegraphics[width=\textwidth]
    {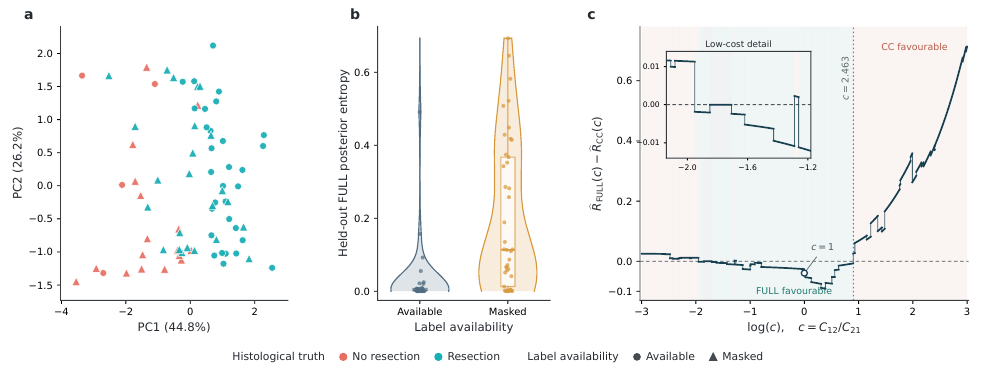}
    \caption{
    Gastrointestinal-lesion application under a shared-covariance
    Gaussian specification.
    Panel~(a) shows the first two standardized principal components
    of the four previously selected features, with histological
    class indicated by colour and label availability by symbol;
    the projection is used only for visualization.
    Panel~(b) shows entropy from the held-out FULL posterior
    predictions.
    Panel~(c) shows the empirical leave-one-out risk difference
    $\widehat R_{\mathrm{FULL}}(c)-\widehat R_{\mathrm{CC}}(c)$.
    Negative values favour FULL and positive values favour CC.
    The inset shows the narrow reversals and tie interval at
    lower cost ratios.
    For the held-out predictions, preprocessing, initialization and fitting
use training observations only, and all 76 paired fits pass the numerical
checks. The principal-component projection in panel~(a) uses all lesions
for visualization only. Cost ratios are sensitivity parameters rather than clinical recommendations.}
    \label{fig:gastro}
\end{figure*}

Panel~(a) of Figure~\ref{fig:gastro} shows the first two standardized
principal components, explaining $44.8\%$ and $26.2\%$ of feature
variation. This projection is used only for visualization.
Panel~(b) compares entropy calculated from the held-out FULL posterior
predictions. Mean entropy is $0.0267$ for lesions with available
labels and $0.2007$ for lesions with masked labels; the corresponding
medians are $0.0024$ and $0.1132$. These descriptive differences do
not establish $M\perp Z\mid Y$ or validate the parametric missingness
model, which remain working assumptions.

Let class~1 denote resection and class~2 no resection.
Set $C_{21}=1$ and $c=C_{12}/C_{21}$, where $C_{12}$ is the loss of
recommending resection for a true no-resection lesion and $C_{21}$
is the loss of failing to recommend resection for a true resection
lesion. For each fitted method, the held-out decision assigns a
lesion to class~1 when
\[
\frac{\widehat\tau_1(y)}{1-\widehat\tau_1(y)}>c.
\]
We hold the fitted posterior predictions fixed and vary only $c$.
At equal costs,
\[
\widehat R_{\mathrm{CC}}(1)=\frac{14}{76}=0.1842,
\qquad
\widehat R_{\mathrm{FULL}}(1)=\frac{11}{76}=0.1447.
\]
Thus FULL makes three fewer observed classification errors.

Panel~(c) displays
\[
\widehat\Delta(c)
=
\widehat R_{\mathrm{FULL}}(c)
-
\widehat R_{\mathrm{CC}}(c)
\]
over $\log c\in[-3,3]$. The risk difference takes both signs:
it is positive at $c=0.1$, negative at $c=1$, and positive at $c=5$.
FULL has lower empirical risk throughout the interval
$0.283236<c<2.462888$, while additional narrow reversals and a tie
interval occur at lower cost ratios. Supplementary Section~\ref{sec:supp-gastro}
reports the complete ordering.

The application illustrates the cost sensitivity of the fitted
decision rules. It does not establish a population predictive
advantage or identify the covariance mechanism derived for the
restricted Gaussian model. The cost ratios are sensitivity
parameters, rather than recommended clinical costs.

\section{Discussion}
\label{sec:discussion}

The paper gives explicit first-order deployment-cost crossings for
the one-dimensional Gaussian discrimination problem and proves that
finite classifier-linked missingness mechanisms can generate the
covariance geometry required for reversal at fixed marginal
missingness. The existence result concerns a likelihood in which the
class-probability logit and separation are estimated, while the
class-mean midpoint and within-class variance are known. Estimating
additional generative parameters requires their information and
nuisance adjustment; canonical standardization alone does not remove
them.

For general finite-action losses, the leading comparison is
$\operatorname{tr}(\boldsymbol H_C\boldsymbol D)$, where
$\boldsymbol D=\boldsymbol V_{\mathrm{PC}}-
\boldsymbol V_{\mathrm{CC}}$ describes estimation uncertainty and
$\boldsymbol H_C$ describes the active decision geometry.
Indefiniteness is sufficient for reversal in the one-dimensional
binary Gaussian case, but need not suffice in higher dimensions or
multiclass problems. The attainable loss-specific curvature matrices
must expose both signs of the covariance difference.

The reported simulation curves display changes in estimated risk
ordering as the deployment loss varies. Their close agreement with
an empirical quadratic comparison is consistent with the local
risk mechanism, including under unequal priors. Differences between
empirical and population crossings can reflect finite-sample effects
and Monte Carlo variability; agreement between two summaries of the
same replications does not separate these contributions.
The gastrointestinal application illustrates cost sensitivity using
agreement-defined availability of histological labels.
The shared-covariance reanalysis uses training-only preprocessing
and initialization, and all $76$ paired fits pass the numerical
checks. FULL has lower observed risk at equal costs, but the
empirical ordering reverses as the deployment cost ratio varies.
This exploratory comparison is conditional on the historical
feature representation; it neither establishes a population
predictive advantage nor identifies the covariance mechanism
derived for the restricted Gaussian model.
The relative costs remain sensitivity parameters rather than
clinical utility estimates.

The theoretical comparison is local and asymptotic, and concerns the
leading risk of regular estimators. It does not establish
finite-sample dominance. Correct specification of the generative
and classifier-linked missingness models is essential to the
information calculation. The multiclass expansion assumes regular
active faces, transverse higher-order ties, integrable boundary
curvature, and uniform quadratic control of regret in the tails.
Positive-measure ties, nontransverse intersections, or uncontrolled
tail decision changes require a different analysis.

Extensions include comparison over a class of uncertain deployment
losses and studying the frontier under classifier or mechanism
misspecification. Designing label acquisition for a specified
deployment loss is a related question. Work on classification-risk-optimal
label acquisition \cite{SetoudehtazangiMcLachlan2026b} concerns changing
the training experiment; the present comparison instead holds that
experiment fixed while varying the deployment loss.

\section*{Statements and Declarations}

\subsection*{Data and code availability}
The gastrointestinal-lesion dataset is publicly available from
the UCI Machine Learning Repository at
\url{https://doi.org/10.24432/C5V02D}.
The analysis inputs, comprising the four previously selected
features, histological reference classes and agreement-defined
label mask, together with the code for the simulations,
real-data analysis and manuscript figures, are available at
\url{https://github.com/fariborz-setoudeh/decision-target-reversal}.

\subsection*{Declaration of competing interest}
The authors declare that they have no known competing
financial interests or personal relationships that could
have appeared to influence the work reported in this paper.
\IEEEtriggeratref{15}
% IEEE numeric references, ordered by first citation.
% Bibliographic content is retained from the supplied manuscript.

\end{document}